\documentclass[runningheads]{llncs}
\usepackage{booktabs} 
\usepackage{diagbox}
\usepackage{xcolor}
\usepackage{graphicx}
\usepackage[colorlinks=true, linkcolor=black, anchorcolor=black, citecolor=black, urlcolor=blue]{hyperref}         

\usepackage{amsmath}
\usepackage{amssymb}
\usepackage{ctable}
\usepackage{graphicx}
\usepackage{mathtools}
\usepackage{multirow}
\usepackage{algorithm}
\usepackage{algpseudocode}
\usepackage{adjustbox}
\usepackage{tabularx}
\usepackage{caption}
\usepackage{float} 
\usepackage{subcaption}
\usepackage{multirow}
\usepackage{colortbl}
\usepackage{float}
\usepackage{tabularx}
\usepackage{soul}
\definecolor{pink}{RGB}{241,167,181}
\definecolor{Green}{RGB}{178,213,155}
\definecolor{Yellow}{RGB}{242,222,121}

\begin{document}
\title{Resource-aware Mixed-precision Quantization for Enhancing Deployability of Transformers for Time-series Forecasting on Embedded FPGAs}
\titlerunning{Resource-Aware Mixed-Precision Quantized Transformers}
\author{Tianheng Ling \and Chao Qian \and Gregor Schiele}
\authorrunning{T. Ling et al.}
\institute{Intelligent Embedded Systems Lab, University of Duisburg-Essen \\ 47057 Duisburg, Germany \\
\email{\{tianheng.ling, chao.qian, gregor.schiele\}@uni-due.de}}
\maketitle    

\begin{abstract}

This study addresses the deployment challenges of integer-only quantized Transformers on resource-constrained embedded FPGAs (Xilinx Spartan-7 XC7S15). We enhanced the flexibility of our VHDL template by introducing a selectable resource type for storing intermediate results across model layers, thereby breaking the deployment bottleneck by utilizing BRAM efficiently. Moreover, we developed a resource-aware mixed-precision quantization approach that enables researchers to explore hardware-level quantization strategies without requiring extensive expertise in Neural Architecture Search. This method provides accurate resource utilization estimates with a precision discrepancy as low as 3\%, compared to actual deployment metrics. Compared to previous work, our approach has successfully facilitated the deployment of model configurations utilizing mixed-precision quantization, thus overcoming the limitations inherent in five previously non-deployable configurations with uniform quantization bitwidths. Consequently, this research enhances the applicability of Transformers in embedded systems, facilitating a broader range of Transformer-powered applications on edge devices.

\end{abstract}

\section{Introduction}
\label{sec:introcution}

The integration of Artificial Intelligence (AI) into the Internet of Things (IoT) is significantly reshaping interactions among devices, humans, and environments, enhancing decision-making capabilities and responsiveness~\cite{chander2022artificial}. Deploying Deep Learning (DL) models directly on edge devices, termed Edge Intelligence, leverages data proximity to improve the timeliness and relevance of computational tasks~\cite{zhou2019edge}. Local data processing under this paradigm can considerably reduce the need for continuous data transmission to the Cloud, thus minimizing latency, conserving bandwidth, and bolstering data privacy and security~\cite{dave2021benefits}.

Despite these advantages, implementing sophisticated DL models, such as Transformers, which are renowned for their efficacy in sequence modeling tasks, including time-series forecasting, encounters significant challenges on IoT devices~\cite{wen2022transformers}. These challenges primarily stem from the limited computational power and memory capacities of such platforms~\cite{chen2020deep}, which are critical for the widespread adoption of autonomous and efficient ubiquitous computing technologies~\cite{gill2022ai}. A notable research gap exists in time-series forecasting with integer-only quantized Transformers on platforms like embedded Field-Programmable Gate Arrays (FPGAs). These FPGAs have been widely used to accelerate DL models but have yet to be thoroughly explored for this specific task ~\cite{seng2021embedded}.

Our previous work~\cite{ling2024transformer} investigated the implementation of integer-only quantized Transformers for time-series forecasting on Spartan-7 XC7S15 FPGA from Xilinx. The results demonstrated that our 4-bit quantized models achieved precision comparable to 6-bit quantized counterparts on a traffic flow dataset~\cite{qian2022enhancing} and even outperformed 8-bit quantized counterparts on an air quality (AirU) dataset~\cite{becnel2022tiny} by 2.83\%. Additionally, we explored the feasibility of deploying models with various configurations on this embedded FPGA, examining their resource utilization, inference time, and power and energy consumption. Our findings indicated that while smaller models could be easily supported, larger models exceeded the resource capacities of the XC7S15 FPGA, thus highlighting a significant scalability issue.

This scalability challenge underpins our current research, which focuses on exploring and enhancing the deployability of Transformers on embedded FPGAs. The primary contributions of this paper are as follows:
\begin{itemize}
    \item We enhanced the configurability of our solution by introducing selectable resource types for storing intermediate results across model layers. This enhancement enables the deployment of two previously infeasible model configurations by mitigating inefficient BRAM utilization.
    
    \item We extended our quantization framework to support mixed-precision quantization for Transformer models, incorporating resource awareness to streamline the selection process for deployable model configurations. By utilizing a knowledge database that catalogs the resource utilization of each model component, we can estimate the resource utilization for each configuration. This predictive capability allows us to pre-select potential candidate models based on threshold and score-based filtering for training and deployment.
    
    \item Our method efficiently navigates the extensive search space without requiring in-depth Neural Architecture Search expertise. The resource estimation based on our knowledge database shows only a 3\% difference compared to Vivado’s estimation, validating the reliability of our database as a robust foundation for advanced techniques, enabling their future integration.

    \item We systematically evaluated our approach against two baseline methods and performed further validation using Vivado’s synthesis and actual hardware measurements. 
    
\end{itemize}

The remainder of this paper is organized as follows: 
Section \ref{sec:transformer} describes the FPGA-friendly Transformer architecture tailored for time-series forecasting.
Section \ref{sec:problem_statement} details the deployment challenges of complex Transformer models on resource-constrained FPGAs. 
Section \ref{sec:solution_design} describes our strategic approaches to overcome these challenges. 
Section \ref{sec:results_evaluation} presents the experimental results validating the effectiveness of our approaches. Section \ref{sec:related_work} reviews the relevant literature. 
Finally, Section \ref{sec:conclusion_future} concludes the paper and suggests avenues for future research.


\section{Transformers for Time-series Forecasting}
\label{sec:transformer}

\begin{figure}[!htb]
    \centering
    \includegraphics[width=1\textwidth]{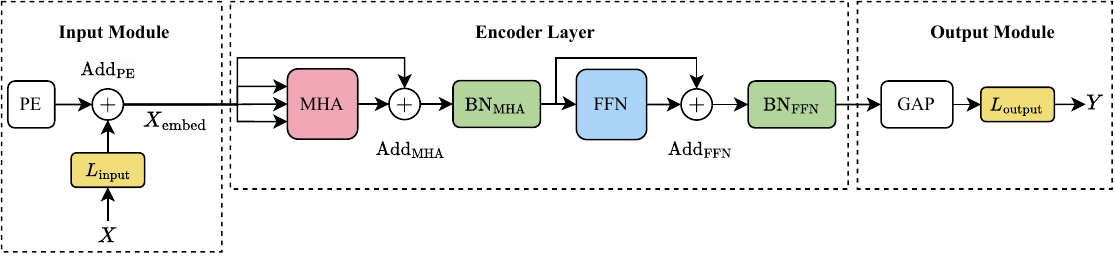}
    \caption{The Architecture of the Transformer Model}
    \label{fig:transformer_model}
\end{figure}

This section revisits the architecture of the FPGA-friendly Transformer tailored for single-step ahead time-series forecasting, as initially proposed in our prior work \cite{ling2024transformer}. Figure~\ref{fig:transformer_model} depicts the model, composed of three primary components: an input module, an encoder layer, and an output module. This modular structure allows the model to handle univariate and multivariate time-series.

\vspace{-5pt}
\vspace{-15pt}

The input module begins by processing the input sequence $X$, consisting of $n$ data points, each with $m$ dimensions. This process is facilitated by a linear layer (denoted $L_\text{input}$), which projects the input data into a higher-dimensional feature space ($d_\text{model}$). Positional Encoding (PE) is applied after that to incorporate positional information into the transformed data, resulting in $X_\text{embed}$ with dimensions $(n, d_\text{model})$. This embedding is then fed into the encoder layer. The encoder layer is structured around a Multi-head Self-attention (MHA) module coupled with a Feedforward Network (FFN) module. Each module is followed by Residual Connection (Add) and Batch Normalization (BN), enhancing stability and learning efficacy. The output module then consolidates the encoder layer's outputs using Global Average Pooling (GAP) and a final linear transformation (denoted $L_\text{output}$) to produce the forecast output $Y$.

We standardized several key dimensions to streamline the architecture and mitigate training complexity. Specifically, the dimensions of the query, key, value, and output vectors within the MHA module were unified at \(d_{\text{model}}\), and the dimension of the FFN was expanded to \(4 \times d_{\text{model}}\). Furthermore, the number of heads \(h\) in the MHA was minimized to 1 for simplicity, as increasing \(h\) does not obviously improve model precision on our target datasets.

In subsequent sections, we will applying mixed-precision quantization on this model by targeting specific key components: $L_\text{input}$, $\text{Add}_\text{PE}$, MHA, $\text{Add}_\text{MHA}$, $\text{BN}_\text{MHA}$, FFN, $\text{Add}_\text{FFN}$, $\text{BN}_\text{FFN}$, GAP, and $L_\text{output}$. These components are selected due to their crucial role in balancing model precision and deployment efficiency. Our design also allows researchers and developers to adapt the granularity by grouping or separating our key components to meet the specific needs of their applications. For example, the FFN module can be dissected into two linear layers with an intervening ReLU activation function. 

\section{Problem Statement}
\label{sec:problem_statement}

Our previous work~\cite{ling2024transformer} evaluated the deployment of Transformer models with various configurations on the XC7S15 FPGA, focusing on different input sequence lengths ($n$), embedding dimensions ($d_{\text{model}}$), and quantization bitwidths ($b$). It highlighted that configurations with smaller dimensions often incurred significant losses in model precision while adhering to hardware constraints. In contrast, bigger configurations had better model precision but frequently exceeded the resource capacities of FPGAs even with low quantization bitwidth.

\vspace{-15pt}
\begin{table}[!htb] 
\centering
\caption{Resource Utilization of Transformers on XC7S15 FPGA~\cite{ling2024transformer}}
\label{tab:model_resource}
\begin{adjustbox}{center}
\begin{tabular}{|c|c|c|c|c|c|c|c|c|c|c|}
\hline
\multirow{3}{*}{\textbf{$n$}} & \multirow{3}{*}{\textbf{$d_\text{model}$}} & \multicolumn{9}{c|}{\textbf{$b$}}\\\cline{3-11} 
 & & \multicolumn{3}{c|}{4} & \multicolumn{3}{c|}{6} & \multicolumn{3}{c|}{8} \\ \cline{3-11} 
 & & \multicolumn{1}{c|}{LUTs} & \multicolumn{1}{c|}{BRAM} & \multicolumn{1}{c|}{DSPs} & \multicolumn{1}{c|}{LUTs} & \multicolumn{1}{c|}{BRAM} & \multicolumn{1}{c|}{DSPs} & \multicolumn{1}{c|}{LUTs} & \multicolumn{1}{c|}{BRAM} & DSPs \\ \hline
\multirow{4}{*}{6}  
&   8 & 34.2 & 10.0 & 65.0 & 42.0 & 10.0 & 90.0 & 55.6 & 10.0 & 100.0  \\ \cline{2-11}
&  16 & 37.3 & 30.0 & 65.0 & 47.0 & 30.0 & 95.0 & 57.1 & 40.0 & 100.0  \\ \cline{2-11}
&  32 & 41.1 & 40.0 & 65.0 & 50.5 & 55.0 & 95.0 & 62.7 & 55.0 & 100.0  \\ \cline{2-11}
&  64 & 47.4 & 75.0 & 60.0 & 57.2 & 100.0 & 90.0 & 89.5 & 100.0 & 100.0 \\ \hline
\multirow{4}{*}{12} 
&   8 & 36.5 & 10.0 & 65.0 & 46.0 & 10.0 & 95.0 & 58.0 & 20.0 & 100.0   \\ \cline{2-11}
&  16 & 42.7 & 35.0 & 60.0 & 51.5 & 35.0 & 95.0 & 65.1 & 45.0 & 100.0 \\ \cline{2-11}
&  32 & 46.8 & 40.0 & 65.0 & 58.9 & 55.0 & 95.0 & 74.3 & 60.0 & 100.0  \\ \cline{2-11}
&  64 & 58.2 & 75.0 & 65.0 & 75.3 & 100.0 & 95.0 & \cellcolor{pink} \textbf{115.0} & \cellcolor{pink} \textbf{100.0} & \cellcolor{pink} \textbf{100.0}  \\ \hline
\multirow{4}{*}{18} 
&   8 & 40.8 & 15.0 & 55.0  & 49.9 & 15.0 & 95.0 & 63.4 & 20.0 & 100.0  \\ \cline{2-11}
&  16 & 45.8 & 35.0 & 65.0  & 55.8 & 35.0 & 95.0 & 71.0 & 45.0 & 100.0  \\ \cline{2-11}
&  32 & 53.0  & 40.0 & 65.0 & 67.0 & 55.0 & 95.0 & 85.3 & 60.0 & 100.0  \\ \cline{2-11}
&  64 & \cellcolor{Green} \textbf{72.8}  &  \cellcolor{Green} \textbf{75.0} & \cellcolor{Green} \textbf{60.0} &  \cellcolor{pink} \textbf{93.6$\dagger$} &  \cellcolor{pink} \textbf{100.0} & \cellcolor{pink} \textbf{90.0} &   \cellcolor{pink} \textbf{136.5} & \cellcolor{pink} \textbf{100.0} & \cellcolor{pink} \textbf{100.0} \\ \hline
\multirow{4}{*}{24} 
&  8  & 39.2 & 15.0 & 50.0 & 52.7 & 20.0 & 90.0 & 67.8 & 20.0 & 100.0  \\ \cline{2-11}
&  16 & 48.7 & 40.0 & 65.0 & 60.7 & 40.0 & 95.0 & 77.4 & 45.0 & 100.0  \\ \cline{2-11}
&  32 & 58.3 & 45.0 & 65.0 & 73.7 & 60.0 & 95.0 & \cellcolor{Yellow} \textbf{99.5$^{\star}$} & \cellcolor{Yellow} \textbf{60.0} & \cellcolor{Yellow} \textbf{100.0} \\ \cline{2-11}
&  64 & \cellcolor{Yellow} \textbf{82.4$^{\star\star}$} & \cellcolor{Yellow} \textbf{80.0} & \cellcolor{Yellow} \textbf{65.0} & \cellcolor{pink} \textbf{107.8} & \cellcolor{pink} \textbf{100.0} & \cellcolor{pink} \textbf{90.0} &   \cellcolor{pink} \textbf{157.6} & \cellcolor{pink} \textbf{100.0} & \cellcolor{pink} \textbf{100.0} 
  \\ \hline
\multicolumn{11}{l}{\small $\dagger$ The utilization of DRAM is 105.5\%.} \\
\multicolumn{11}{l}{\small $\star$ The utilization of DRAM is 112.5\%.} \\
\multicolumn{11}{l}{\small $\star\star$ The utilization of DRAM is 104.4\%.} 
\end{tabular}
\end{adjustbox}
\end{table}
\vspace{-10pt}

To gain a detailed understanding of the specific resource bottlenecks that prevented these model configurations from being deployed, we executed multiple Vivdao syntheses for these eight configurations, whose resource utilization ware highlighted in the Table \ref{tab:model_resource}. The 4-bit quantized model with configuration \(n\!=\!18\) and \(d_\text{model}\!=\!64\) was previously considered too resource-intensive to be deployed. As colored in green in Table \ref{tab:model_resource}, this model now meets FPGA constraints with additional synthesis iterations. This discrepancy indicates that local minima might have affected initial synthesis, highlighting the need for multiple synthesis iterations to verify deployability accurately.

Further investigation revealed that two configurations (colored in yellow in Table \ref{tab:model_resource}) maxed out LUTs as memory (DRAM) but left significant portions of Block RAM (BRAM) underutilized. For instance, the 8-bit quantized model with \(n\!=\!24\) and \(d_\text{model}\!=\!32\) displayed near-maximal utilization of Look-Up Tables (LUTs) at 99.5\% and DRAM at 112.5\%, yet 40\% of BRAM remained unused. Similarly, the 4-bit model with \(n\!=\!24\) and \(d_\text{model}\!=\!64\) utilized 82.4\% of LUTs and 104.42\% of DRAM, with 20\% of BRAM still available. These findings indicate that prior implementations may over-allocate DRAM resources while leaving a significant portion of BRAM underutilized. Thus, an in-depth investigation into optimizing resource allocation is warranted to enhance efficiency.

Moreover, as colored in pink in Table \ref{tab:model_resource}, the remaining five configurations were non-deployable as they exhausted all FPGA resources. For example, the 8-bit quantized model with $n\!=\!12$ and $d_\text{model}\!=\!64$ exceeded the FPGA’s LUT capacity and fully utilized the available BRAM, rendering deployment infeasible. Similarly, while the LUT utilization for the 6-bit quantized model with $n\!=\!18$ and $d_\text{model}\!=\!64$ remained within acceptable limits, its DRAM was over-utilized at 105.5\%.

We believe that quantization with uniform bitwidth across all model layers may be contributing to excessive resource consumption and unacceptable drops in precision. To address these issues, recent studies~\cite{nagel2021white,rakka2024review,ding20224} advocate for mixed-precision quantization, which assigns varied bitwidths to different layers or components. However, this approach necessitates the efficient selection of optimal combinations of layer-specific bitwidths. The entire workflow—from selecting quantization bitwidths and conducting quantization-aware training (QAT) to model convergence, synthesizing hardware accelerators, and reviewing Vivado reports—often results in substantial time and effort. This issue is particularly pronounced when, after extensive efforts, the final model fails to fit the FPGA~\cite{dong2020hawq}. Thus, integrating resource awareness early in the workflow is crucial to streamline the deployment process and mitigate potential inefficiencies.

\section{Proposed Solutions}
\label{sec:solution_design}

This section delineates our proposed solutions for embedded FPGAs. Building on the challenges identified in Section~\ref{sec:problem_statement}, we focus on two principal approaches: 1) adaptive resource allocation and 2) resource-aware mixed-precision quantization. 

\subsection{Adaptive Resource Allocation}

In our prior designs of the Transformer accelerator~\cite{ling2024transformer}, memory allocation was predominantly designated for two purposes: 1) storing model parameters (including weights, biases, and positional encoding information) in BRAM, and 2) accommodating intermediate computational results in DRAM. Initially, it was presumed that model parameters would necessitate more memory than intermediate results, leading to the prioritization of DRAM for intermediate results to maximize the availability of BRAM for model parameters. However, subsequent analysis of accelerator implementations from our earlier work~\cite{ling2024transformer} indicated that this approach could lead to underutilization of BRAM resources and overutilization of DRAM, as mentioned in Section~\ref{sec:problem_statement}. This issue becomes more pronounced with increasing input sequence lengths in the Transformer model, where the volume of intermediate computation results can surpass that of the model parameters.

To rectify this misalignment, we introduce a configurable parameter within the VHDL template that explicitly specifies the type of resource: 1) BRAM, 2) DRAM, or 3) automatic allocation (BRAM or DRAM) by Vivado for storing intermediate results. This enhancement introduces flexibility in resource allocation for intermediate results, allowing adaptations based on each model's specific condition. For instance, intermediate results can be organized by size at the model level, prioritizing the storage of larger intermediate results in BRAM wherever feasible, thereby optimizing resource efficiency. In this work, we prefer to apply option 3 to rely on Vivado's algorithm to find feasible solutions. Manual intervention remains possible, enabling users to adjust settings if automatic allocation does not yield the desired outcomes. With option 3, we successfully deployed two models previously deemed non-deployable, highlighted in yellow in Table~\ref{tab:model_resource}, on the XC7S15 FPGA. Detailed results of these deployments are thoroughly discussed in Section~\ref{subsec:experiment1}.

\subsection{Mixed-precision Quantized Transformer}

Building upon the optimized resource allocation, we further explore the resource-aware mixed-precision quantization. This exploration unfolds in two steps: 1) transitioning from uniform to mixed-precision quantization and 2) introducing resource awareness within the mixed-precision quantization.

\subsubsection{Transition to Mixed-precision Quantization}

\begin{figure}[!htb]
    \centering
    \includegraphics[width=.9\textwidth]{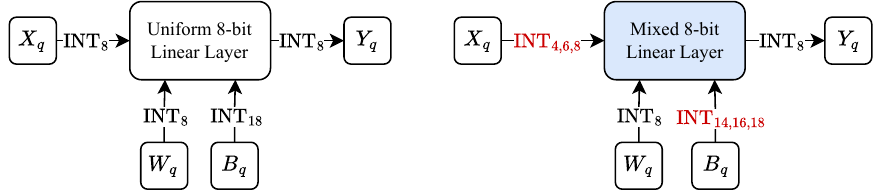}
    \caption{Uniform (left) vs Mixed (right) 8-bit Linear Layer}
    \label{fig:mixed_linear}
\end{figure}

In our preceding study~\cite{ling2024transformer}, quantization with uniform bitwidths was applied across all layers. As detailed on the left side in Figure~\ref{fig:mixed_linear}, in a typical linear layer under uniform 8-bit quantization, inputs (\(X_{q}\)), outputs (\(Y_{q}\)), and weights (\(W_{q}\)) were processed using an 8-bit asymmetric scheme. To reduce computational overhead, biases (\(B_{q}\)) adopted a symmetric quantization scheme using a quantization scale factor determined by \(X_{q}\) and \(W_{q}\). The quantization bitwidth for \(B_{q}\) was set to 18 bits, a value derived from the bitwidth of the multiply-accumulator unit in this linear layer.

The current work expands on these foundations and introduces a mixed-precision quantization, permitting individual layers to operate with distinct bitwidths. In our case, the quantization bitwidth of the layer's inputs is dictated by the quantization bitwidth of the preceding layer's outputs to avoid additional rescaling caused by bitwidth mismatches. Hence, it is not specified independently within our naming conventions. Illustratively, as depicted on the right side of Figure~\ref{fig:mixed_linear}, using mixed 8-bit quantization as example, where \(W_{q}\) and \(Y_{q}\) within a linear layer are quantized at 8 bits, while the \(X_{q}\) can vary between 4, 6, or 8 bits depending on the preceding layer's \(Y_{q}\). This cascading effect also influences the quantization bitwidth of \(B_{q}\), which varies from 14 to 18 bits.

\vspace{-10pt}
  \begin{figure}[!htb]
    \centering
    \includegraphics[width=.9\textwidth]{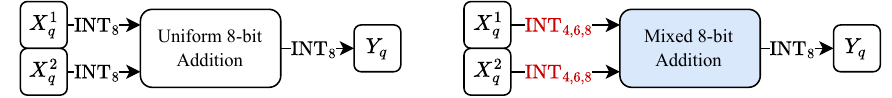}
    \caption{Uniform (left) vs Mixed (right) 8-bit Addition Operation}
    \label{fig:mixed_add}
\end{figure}
\begin{figure}[!htb]
    \centering
    \includegraphics[width=.9\textwidth]{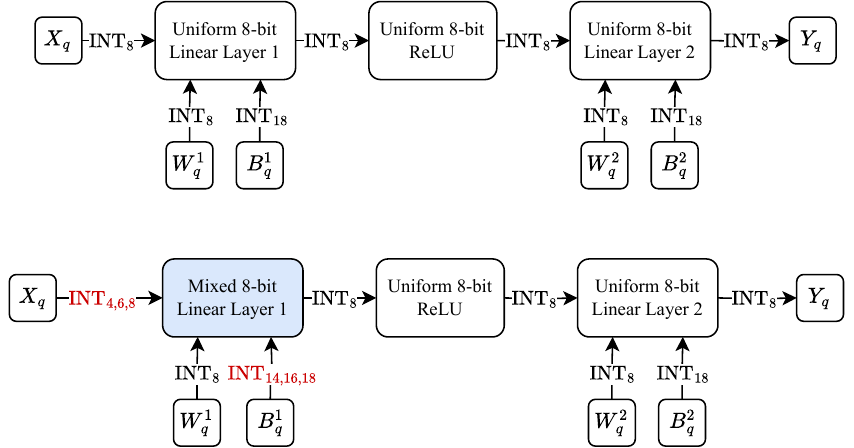}
    \caption{Uniform (up) vs Mixed (down) 8-bit FFN Module}
    \label{fig:mixed_ffn}
\vspace{-10pt}
\end{figure}


This principle of mixed-precision quantization are also applied to other operations and modules in the Transformer model. For instance, Figure~\ref{fig:mixed_add} illustrates a mixed 8-bit quantized addition operation where the bitwidths of its Inputs1 (\(X^{1}_{q}\)) and Inputs2 (\(X^{2}_{q}\)) are determined by the previous layers' outputs, but its outputs (\(Y_{q}\)) remains at 8 bits. For mixed-precision quantized modules such as the FFN, depicted in Figure~\ref{fig:mixed_ffn}, only the initial linear layer utilizes mixed 8-bit quantization. Subsequent operations within this module consistently maintain uniform 8-bit quantization.

\subsubsection{Resource-aware Mixed-precision Quantization}

\begin{figure}[!htb]
    \centering
    \includegraphics[width=.9\textwidth]{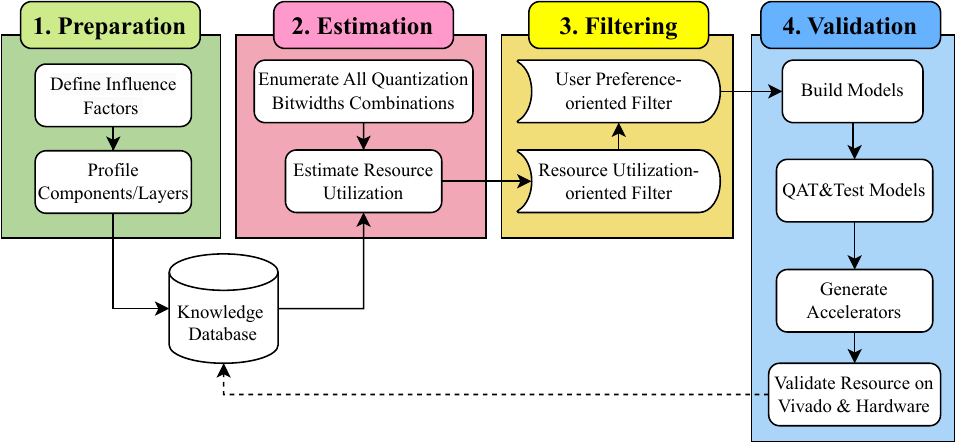}
    \caption{Workflow of Resource-aware Mixed-precision Quantization}
    \label{fig:workflow}
    \vspace{-10pt}
\end{figure}

As illustrated in Figure \ref{fig:workflow}, we proposed an extended workflow of implementing our resource-aware approach to the mixed-precision quantized Transformers. This workflow is segmented into four phases:  1) Preparation, 2) Estimation, 3) Filtering, and 4) Validation. Each phase plays a crucial role in ensuring that the models are optimized for the resource constraints of embedded FPGAs.

\paragraph{Phase 1: Preparation}

The initial phase involves developing a comprehensive knowledge database for subsequent phases. Initially, we identify key factors influencing model precision and resource utilization. We then select the key components of the model for profiling. Following this, variations in these factors are systematically explored. This process involves training models under varying conditions, generating accelerators, and synthesizing detailed resource utilization reports. Each report details the resource breakdown for individual components, culminating in a comprehensive lookup table called the knowledge database. The preparation of this database will be detailed in \ref{subsec:knowledge_database}.

\paragraph{Phase 2: Estimation}

Utilizing the knowledge database, this phase predicts the resource utilization of mixed-precision quantized models with various quantization bitwidth combinations. The estimation process entails consulting the knowledge database for each key component to assess its resource consumption. These individual estimations are then aggregated to ascertain the total resource consumption of a model. Users can enumerate all quantization bitwidth combinations or specify a subset based on their domain knowledge and specific requirements. In Section \ref{subsec:res-filter}, we will further introduce the resource estimation process executed in our study.

\paragraph{Phase 3: Filtering}

The filtering phase strategically selects a limited number of model candidates for further training and accelerator generation. Initially, this phase employs a threshold-based filter for each resource type, which excludes models whose estimated resource consumption exceeds predefined limits. It can be supplemented by additional user-defined criteria that refine the selection process to meet specific deployment requirements. Section \ref{subsec:res-filter} will explain the filters implemented in our study.

\paragraph{Phase 4: Validation}

The final phase validates models with the chosen bitwidth combinations through a rigorous process involving model construction, QAT, and performance testing. Each model’s resource utilization and operational efficiency are evaluated using Vivado and actual FPGA hardware. In addition, while not yet implemented, reports generated during the validation phase hold the potential to enrich the knowledge database, thereby enhancing its predictive accuracy and robustness.

\subsection{Preparation of Knowledge Database}
\label{subsec:knowledge_database}

To address the deployability challenges identified in prior research \cite{ling2024transformer}, this study defines three influence factors for constructing a knowledge database: 1) input sequence length ($n$), 2) embedding dimension ($d_{\text{model}}$), and 3) quantization bitwidth ($b$). Expressly, $d_{\text{model}}$ is uniformly set to 64, while sequence lengths are selected as 12, 18, and 24, paired with uniform quantization bitwidths of 4, 6, and 8 bits. This setup creates a total of 9 different model configurations.

As mentioned in Section~\ref{sec:transformer}, our study mainly profiles model components: $L_\text{input}$, $\text{Add}_\text{PE}$, MHA, $\text{Add}_\text{MHA}$, $\text{BN}_\text{MHA}$, FFN, $\text{Add}_\text{FFN}$, $\text{BN}_\text{FFN}$, GAP, and $L_\text{output}$. Furthermore, our study also incorporates the resource overhead from $O_\text{model}$, $O_\text{encoder layer}$, and $O_\text{middleware}$, attributable to LUTs required for component interconnections and data buffering.

\begin{figure}[!htb]
\vspace{-10pt}
    \centering
    \includegraphics[width=1\textwidth]{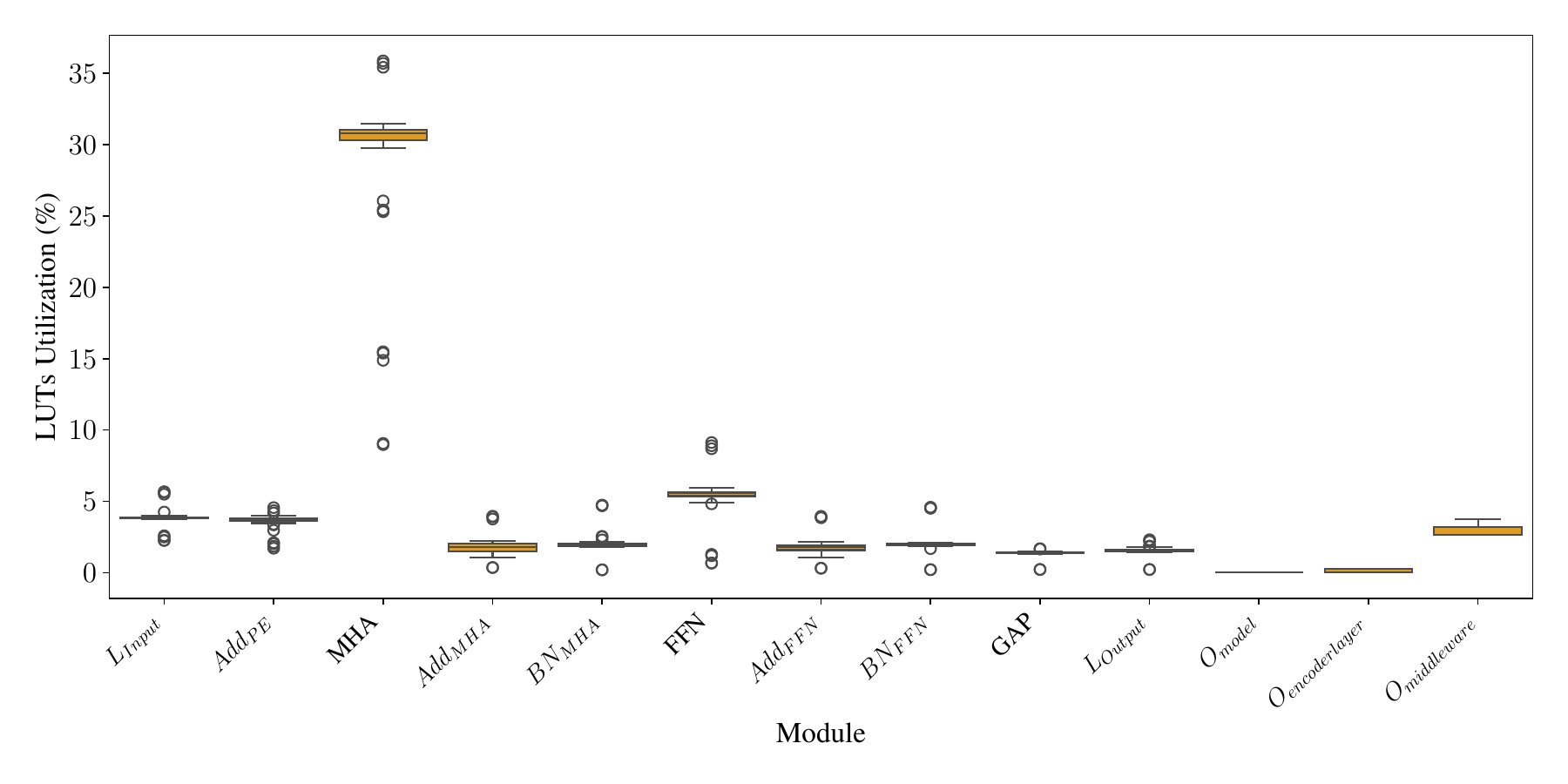}
    \caption{LUTs Utilization Cross Key Components for $n\!=\!12$, $d_{\text{model}}=64$ and $b\!=\!4$}
    \label{fig:boxplot_luts_12_64}
\vspace{-10pt}
\end{figure}

For each model configuration, 50 training sessions are conducted, generating 450 distinct accelerators. Each accelerator undergoes five separate synthesis iterations to aggregate the knowledge database, primarily focusing on LUTs, DRAM, BRAM, and DSPs. Derived from 250 resource utilization reports, Figure \ref{fig:boxplot_luts_12_64} demonstrates the distribution of LUTs utilization across key components, focusing on the configuration where $n\!=\!12$ and \(b\!=\!4\). The analysis reveals that the LUTs utilization of each component has a slight variation. However, the presence of outliers, likely due to the optimization algorithms utilized by Vivado, suggests variability in synthesis outcomes. 

To mitigate the impact of such variability, median values are employed to provide a more stable representation of resource utilization. Through an extensive analysis of 2250 utilization reports, we constructed a knowledge database, as cataloged in Table \ref{table:knowledge_database}. This database has been systematically validated for accuracy and dependability on quantized models with uniform quantization bitwidths. For instance, the median LUT utilization for the model configuration with \(n\!=\!12\) at a 4-bit quantization is predicted to be 57.2\%, aligning closely with previous measurements of 58.1\% on actual FPGA hardware (as detailed in Table \ref{tab:model_resource}), confirming the reliability of the database.

\begin{table}[!htbp]
\footnotesize
\vspace{-10pt}
\centering
\caption{Knowledge Database of Resource Utilization}
\label{table:knowledge_database}
\begin{tabular}{|c|c|c|c|c|c|c|c|c|c|c|c|c|c|}
\hline
\multirow{2}{*}{$n$} & \multirow{2}{*}{Components}  & \multicolumn{3}{c|}{LUTs} & \multicolumn{3}{c|}{DRAM} & \multicolumn{3}{c|}{BRAM} & \multicolumn{3}{c|}{DSPs} \\ \cline{3-14} 

& & \multicolumn{1}{c|}{4-bit} & \multicolumn{1}{c|}{6-bit} & 8-bit & \multicolumn{1}{c|}{4-bit} & \multicolumn{1}{c|}{6-bit} & 8-bit & \multicolumn{1}{c|}{4-bit} & \multicolumn{1}{c|}{6-bit} & 8-bit & \multicolumn{1}{c|}{4-bit} & \multicolumn{1}{c|}{6-bit} & 8-bit \\ \hline

\multirow{13}{*}{12}  & $L_\text{input}$    & 3.9   & 5.6   & 7.1   & 4.0   & 5.3   & 8.0   & 5.0 & 0.0 & 0.0 & 5.0   & 5.0   & 5.0   \\ \cline{2-14} 
& $\text{Add}_\text{PE}$  & 3.8   & 4.4   & 5.7   & 4.0   & 5.3   & 8.0   & 5.0 & 0.0 & 0.0 & 0.0   & 10.0  & 10.0  \\ \cline{2-14} 
& MHA       & 30.8  & 35.6  & 61.3  & 14.3  & 29.8  & 44.7  & 15.0 &  30.0 &  10.0 & 30.0  & 30.0  & 30.0  \\ \cline{2-14} 
& $\text{Add}_\text{MHA}$  & 1.8   & 3.9   & 5.1   & 0.0   & 5.3   & 8.0   & 5.0 & 0.0 & 0.0 & 0.0   & 10.0  & 10.0  \\ \cline{2-14} 
& $\text{BN}_\text{MHA}$     & 1.9   & 4.7   & 5.3   & 0.0   & 5.3   & 8.0   & 5.0 & 0.0 & 0.0 & 5.0   & 5.0   & 10.0  \\ \cline{2-14} 
& FFN        & 5.4   & 8.9   & 11.4  & 0.0   & 5.3   & 8.0   & 55.0  & 70.0   & 90.0 & 10.0  & 10.0  & 10.0  \\ \cline{2-14} 
& $\text{Add}_\text{FFN}$  & 1.8   & 3.9   & 5.0   & 0.0   & 5.3   & 8.0   & 5.0 & 0.0 & 0.0 & 0.0   & 10.0  & 10.0  \\ \cline{2-14} 
& $\text{BN}_\text{FFN}$     & 2.0   & 4.5   & 5.0   & 0.0   & 5.3   & 8.0   & 5.0 & 0.0 & 0.0 & 5.0   & 5.0   & 10.0  \\ \cline{2-14} 
& GAP      & 1.4   & 1.6   & 1.9   & 0.3   & 0.3   & 0.5   & 0.0 & 0.0 & 0.0 & 5.0   & 5.0   & 5.0   \\ \cline{2-14} 
& $L_\text{output}$    & 1.8   & 2.2   & 2.4   & 0.2   & 0.2   & 0.3   & 0.0 & 0.0 & 0.0 & 5.0   & 5.0   & 5.0   \\ \cline{2-14} 
& $O_\text{model}$   & 0.0   & 0.0   & 0.0   & 0.0   & 0.0   & 0.0 & 0.0 & 0.0 & 0.0   & 0.0   & 0.0   & 0.0   \\ \cline{2-14} 
& $O_\text{encoder layer}$ & 0.0   & 0.3   & 0.3   & 0.0   & 0.0   & 0.0 & 0.0 & 0.0 & 0.0   & 0.0   & 0.0   & 0.0   \\ \cline{2-14} 
& $O_\text{middleware}$         & 2.7   & 3.2   & 3.7   & 0.5   & 0.7   & 0.0 & 0.0 & 0.0 & 1.0   & 0.0   & 0.0   & 0.0   \\ \hline

\multirow{13}{*}{18}  & $L_\text{input}$    & 5.6   & 8.1   & 10.4  & 8.0  & 10.7  & 16.0  & 5.0 & 0.0 & 0.0 & 5.0    & 5.0    & 5.0         \\  \cline{2-14} 
& $\text{Add}_\text{PE}$  & 5.4   & 6.7   & 8.9   & 8.0  & 10.7  & 16.0  & 5.0 & 0.0 & 0.0 & 0.0    & 10.0   & 10.0        \\ \cline{2-14} 
& MHA        & 40.1  & 53.4  & 86.1  & 36.3 & 59.2  & 88.7  & 10.0 & 20.0 &   0.0 & 0.0   & 30.0   & 30.0     \\ \cline{2-14} 
& $\text{Add}_\text{MHA}$  & 1.8   & 6.2   & 8.4   & 0.0  & 10.7  & 16.0  & 5.0 & 0.0 & 0.0 & 0.0    & 10.0   & 10.0        \\ \cline{2-14} 
& $\text{BN}_\text{MHA}$     & 1.9   & 6.9   & 8.3   & 0.0  & 10.7  & 16.0  & 5.0 & 0.0 & 0.0 & 5.0    & 5.0    & 10.0        \\ \cline{2-14} 
& FFN        & 5.6   & 11.4  & 14.6  & 0.0  & 10.7  & 16.0  & 60.0 & 75.0 & 100.0 & 10.0   & 10.0   & 10.0    \\ \cline{2-14} 
& $\text{Add}_\text{FFN}$  & 1.8   & 6.2   & 8.4   & 0.0  & 10.7  & 16.0  & 5.0 & 0.0 & 0.0 & 0.0    & 10.0   & 10.0        \\ \cline{2-14} 
& $\text{BN}_\text{FFN}$     & 1.9   & 6.7   & 8.2   & 0.0  & 10.7  & 16.0  & 5.0 & 0.0 & 0.0 & 5.0    & 5.0    & 10.0        \\ \cline{2-14} 
& GAP      & 1.4   & 1.8   & 2.0   & 0.3  & 0.3   & 0.5   & 0.0 & 0.0 & 0.0 & 5.0    & 5.0    & 5.0         \\ \cline{2-14} 
& $L_\text{output}$    & 1.6   & 2.2   & 2.4   & 0.2  & 0.2   & 0.3   & 0.0 & 0.0 & 0.0 & 5.0    & 5.0    & 5.0         \\ \cline{2-14} 
& $O_\text{model}$   & 0.0   & 0.0   & 0.0  & 0.0   & 0.0   & 0.0 & 0.0 & 0.0 & 0.0    & 0.0    & 0.0    & 0.0    \\ \cline{2-14} 
& $O_\text{encoder layer}$ & 0.3   & 0.4   & 0.6  & 0.0   & 0.0   & 0.0 & 0.0 & 0.0 & 0.0    & 0.0    & 0.0    & 0.0    \\ \cline{2-14} 
& $O_\text{middleware}$ & 3.0   & 3.4   & 3.7   & 0.5  & 0.7   & 1.0   & 0.0 & 0.0 & 0.0 & 0.0    & 0.0    & 0.0         \\ \hline

\multirow{13}{*}{24}  & $L_\text{input}$    & 5.5  & 7.8   & 10.4  & 8.0  & 10.7  & 16.0  & 5.0 & 0.0 & 0.0 & 5.0    & 5.0    & 5.0    \\ \cline{2-14} 
& $\text{Add}_\text{PE}$  & 5.4  & 6.7   & 8.9   & 8.0  & 10.7  & 16.0 & 5.0 & 0.0 & 0.0  & 0.0    & 10.0   & 10.0   \\ \cline{2-14} 
& MHA        & 45.2 & 55.6  & 89.2  & 48.3 & 64.5  & 96.7  & 5.0 & 20.0  & 0.0 &30.0   & 30.0   & 30.0   \\ \cline{2-14} 
& $\text{Add}_\text{MHA}$  & 2.0  & 6.2   & 8.4   & 0.0  & 10.7  & 16.0  & 5.0 & 0.0 & 0.0 & 0.0    & 10.0   & 10.0   \\ \cline{2-14} 
& $\text{BN}_\text{MHA}$     & 2.0  & 6.9   & 8.3   & 0.0  & 10.7  & 16.0  & 5.0 & 0.0 & 0.0 & 5.0    & 5.0    & 10.0   \\ \cline{2-14} 
& FFN        & 5.6  & 11.3  & 14.6  & 0.0  & 10.7  & 16.0  & 60.0 & 75.0  & 100.0 & 10.0   & 10.0   & 10.0   \\ \cline{2-14} 
& $\text{Add}_\text{FFN}$  & 1.8  & 6.2   & 8.3   & 0.0  & 10.7  & 16.0  & 5.0 & 0.0 & 0.0 & 0.0    & 10.0   & 10.0   \\ \cline{2-14} 
& $\text{BN}_\text{FFN}$     & 1.9  & 6.8   & 8.2   & 0.0  & 10.7  & 16.0  & 5.0 & 0.0 & 0.0 & 5.0    & 5.0    & 10.0   \\ \cline{2-14} 
& GAP      & 1.4  & 1.7   & 2.0   & 0.3  & 0.3   & 0.5   & 0.0   & 0.0   & 0.0& 5.0    & 5.0    & 5.0    \\ \cline{2-14} 
& $L_\text{output}$    & 1.6  & 2.3   & 2.4   & 0.2  & 0.2   & 0.3   & 0.0   & 0.0   & 0.0 & 5.0    & 5.0    & 5.0    \\ \cline{2-14} 
& $O_\text{model}$ & 0.0  & 0.0   & 0.0   & 0.0  & 0.0   & 0.0  & 0.0   & 0.0   & 0.0 & 0.0    & 0.0    & 0.0    \\ \cline{2-14} 
& $O_\text{encoder layer}$ & 0.0  & 0.4   & 0.6   & 0.0  & 0.0   & 0.0   & 0.0   & 0.0 & 0.0   & 0.0    & 0.0    & 0.0    \\ \cline{2-14} 
& $O_\text{middleware}$  & 2.7  & 3.3   & 3.9   & 0.0  & 1.0   & 1.5   & 5.0 & 0.0 & 0.0 & 0.0    & 0.0    & 0.0    \\ \hline

\end{tabular}
\vspace{-21pt}
\end{table}

\subsection{Resource Estimation and Filtering}
\label{subsec:res-filter}

In this study, we introduce an algorithm designed to perform resource estimation and filtering efficiently. Algorithm \ref{alg:resource_aware_quantization} supports the systematic selection of mixed-precision quantization bitwidth combinations. This algorithm commences with selecting input sequence lengths ($n$), correlated with the entries in the knowledge database $K_n$. Furthermore, upper limit thresholds for each resource type, including $T_{\text{LUTs}}$, $T_{\text{DRAM}}$, $T_{\text{BRAM}}$, and $T_{\text{DSPs}}$, are established to define the maximum permissible resource usage. A set of quantization bitwidth combinations ($C_{\text{all}}$) that the user wants to explore should also be input.

\begin{algorithm}[!htbp]
\footnotesize
\caption{The Process of Selecting Mixed-precision Combination}
\label{alg:resource_aware_quantization}
\begin{algorithmic}[1]
\State \textbf{Input:} Knowledge database $K_n$, thresholds $T_\text{LUTs}$, $T_\text{DRAM}$, $T_\text{BRAM}$, $T_\text{DSPs}$, quantization combinations $C_\text{all}$
\State \textbf{Output:} Selected quantization combinations $C_\text{selected}$
\State \textbf{Procedure:}
\State Initialize an empty list $C_\text{filtered}$
\For{each combination $c$ in $C_\text{all}$}
    \State Estimate $\text{Util}_\text{LUTs}$, $\text{Util}_\text{DRAM}$, $\text{Util}_\text{BRAM}$, $\text{Util}_\text{DSPs}$ for $c$ using $K_n$
    \If{$\text{Util}_\text{LUTs}  \leq T_\text{LUTs}$ \textbf{and} $\text{Util}_\text{DRAM}  \leq T_\text{DRAM}$ \textbf{and} $\text{Util}_\text{BRAM}  \leq T_\text{BRAM}$ \textbf{and} $\text{Util}_\text{DSPs}  \leq T_\text{DSPs}$}
        \State Compute $sum(c) = \sum_{i=1}^{10} c_i$
        \State Append $[c, sum(c)]$ to $C_\text{filtered}$
    \EndIf
\EndFor
\State Sort $C_\text{filtered}$ by $sum(c)$ in descending order
\State $C_\text{selected} \leftarrow \text{Select the top 5 combinations from } C_\text{filtered}$
\State \textbf{Output:} $C_\text{selected}$
\end{algorithmic}
\end{algorithm}

Initially, the resource utilization for each combination is estimated and compared against these thresholds, as detailed in steps $5\!-\!11$ of Algorithm \ref{alg:resource_aware_quantization}. Combinations that adhere to these criteria are then included in $C_{\text{filtered}}$. Applying a higher threshold will potentially lead to the remaining combinations having more components quantized to higher bitwidth. Decreasing these thresholds allows us to spare resources for other logic, such as data preprocessing when necessary.

Despite resource constraint filtering, the number of feasible combinations can remain extensive. To manage this effectively, we applied sorting on them, outlined in lines 8, 9, and 12 of the algorithm, where combinations $C_\text{filtered}$ are evaluated based on the cumulative sum of their layers' quantization bitwidths. These scores are subsequently used to rank the combinations, selecting the top five $C_\text{selected}$ for further development into mixed-precision quantized models. This scoring mechanism, although fundamental, lays the groundwork for more advanced filtering techniques in future enhancements. Notably, our algorithm identifies five viable candidates of 59,049 combinations within 10 seconds for each $n$, whereas in previous work, synthesizing a single accelerator with Vivado required over one minute.

\section{Experiments and Evaluation}
\label{sec:results_evaluation}

This section describes the experimental setup and subsequent analyses conducted to validate the efficacy of the proposed adaptive resource allocation and our resource-aware mixed-precision quantization. 

\subsection{Experiments Setup}

Out experiments utilize the \textit{AirU} dataset\footnote{\UrlFont{https://dx.doi.org/10.21227/aeh2-a413}}, which comprises multivariate air quality measurements with 19,380 data entries. After rectifying discontinuities, the dataset was narrowed to 15,258 feature-target pairs, allocated into 14,427 for training and 831 for testing, aligning with the test period configuration reported in \cite{becnel2022tiny}. Data normalization was uniformly applied using MinMax scaling to ensure consistency of input features.

Each model configuration underwent 50 training sessions, with each session comprising 100 epochs. An early stopping mechanism with 10 patience epochs was employed to mitigate overfitting. The training was executed using batches of 256 samples, with the Adam optimizer (with $\beta_1\!=\!0.9$, $\beta_2\!=\!0.98$, $\epsilon\!=\!10^{-9}$). The learning rate was set at 0.001 and was adjusted through a decay scheduler, which halved the rate every three epochs. These training sessions were conducted on an NVIDIA GeForce RTX 2080 SUPER GPU, utilizing CUDA 11.0 and PyTorch 3.11 within the Ubuntu operating system. The objective metric for training was the minimization of the Mean Squared Error. In evaluation, model outputs and targets were inverse transformed from their normalized states, and the Root Mean Square Error (RMSE) was computed to assess model performance.

The quantized models were converted into their hardware equivalents through Python scripts that translated model and quantization parameters into VHDL code using predefined templates. The VHDL files were synthesized in Vivado to evaluate resource utilization, timing, and power metrics. The performance and efficiency of the synthesized FPGA accelerators were validated on the ElasticNode V5 hardware platform~\cite{qian2023elasticai}, equipped with an XC7S15 FPGA.

\subsection{Experiment 1: Adaptive Resource Allocation}
\label{subsec:experiment1}

This experiment was designed to assess the efficacy of the adaptive resource allocation approach on XC7S15 FPGA. By employing optimized VHDL templates and configuring the storage of intermediate results with the automatic allocation option, we allowed Vivado to optimize resource allocation within the FPGA constraints autonomously. Expressly, the synthesis strategy was set to maximize BRAM utilization, anticipating full utilization prior to the saturation of DRAM.

\vspace{-20pt}
\begin{table}[]
\centering
\caption{Optimized Resource Utilization of Two Model Configurations}
\label{tab:model_resource_optimized}
\setlength{\tabcolsep}{4 mm}{ 
\begin{tabular}{|clc|cccc|}
\hline
\multicolumn{3}{|c|}{\multirow{2}{*}{\begin{tabular}[c]{@{}c@{}}Configs.\\ ($n$,$d_\text{model}$,$b$)\end{tabular}}} & \multicolumn{4}{c|}{Resource Utilization (\%)} \\ \cline{4-7} 
\multicolumn{3}{|c|}{} & \multicolumn{1}{c|}{LUTs} & \multicolumn{1}{c|}{DRAM} & \multicolumn{1}{c|}{BRAM} & DSPs \\ \hline
\multicolumn{2}{|c|}{\multirow{2}{*}{(24, 32, 8)}}  
& in \cite{ling2024transformer} & \multicolumn{1}{c|}{99.5}     & \multicolumn{1}{c|}{112.5}     & \multicolumn{1}{c|}{60.0}     &  100    \\ \cline{3-7} 
\multicolumn{2}{|c|}{}  & this work  & \multicolumn{1}{c|}{73.3}      & \multicolumn{1}{c|}{41.3}     & \multicolumn{1}{c|}{100.0}   & 100.0       \\ \hline

\multicolumn{2}{|c|}{\multirow{2}{*}{(24, 64, 4)}}  & in \cite{ling2024transformer}                        & \multicolumn{1}{c|}{82.4}    & \multicolumn{1}{c|}{104.4}     & \multicolumn{1}{c|}{80.0}     &   65.0  \\ \cline{3-7} 
\multicolumn{2}{|c|}{}    & this work &  \multicolumn{1}{c|}{74.6}      & \multicolumn{1}{c|}{56.8}    & \multicolumn{1}{c|}{100.0}         & 65.0      \\ \hline
\end{tabular}
}
\end{table}
\vspace{-10pt}

Table~\ref{tab:model_resource_optimized} shows the current resource utilization for model configurations that previously exceeded FPGA constraints (highlighted in yellow in Table~\ref{tab:model_resource}). These configurations now conform to fit the XC7S15 FPGA. Notably, BRAM utilization reached 100\%, affirming our synthesis strategy's efficacy. 

\subsection{Experiment 2: Resource-aware Mixed-precision Quantization}
\label{subsec:experiment2}

Based on the optimization in Experiment 1, this experiment assesses the feasibility of our resource-aware mixed-precision quantization. Focusing on the ten key components identified in Section \ref{sec:transformer}, each component has the option of being quantized at 4, 6, or 8 bits. This setup generates a total of $3^{10}\!=\!59,049$ potential combinations for each input sequence length ($n$). Notably, components such as $O_\text{model}$, $O_\text{encoder layer}$, and $O_\text{middleware}$ are automatically adjusted based on the key components' settings to streamline the deployment process. Moreover, to verify the effectiveness of our approach, we compare it against two baselines.

\subsubsection{Baseline 1: Random Combination}
\label{subsubsec:baseline1}

We randomly selected five quantization bitwidth combinations from 59,049 potential mixed-precision configurations to establish Baseline 1 and constructed models for three distinct input sequence lengths ($n$), totaling 15 configurations. Table \ref{table:combinations_1} describes the experimental outcomes, showcasing the minimum RMSE and the median resource utilization for each configuration across FPGA resources. The ``combinations'' column specifies the bitwidth combination for each component: $L_\text{input}$, $\text{Add}_\text{PE}$, MHA, $\text{Add}_\text{MHA}$, $\text{BN}_{MHA}$, FFN, $\text{Add}_\text{FFN}$, $\text{BN}_{FFN}$, GAP, and $L_\text{output}$.

As highlighted in pink in Table~\ref{table:combinations_1}, we found that 10 of the 15 configurations failed to meet FPGA resource constraints. We assume that the high failure rate in model deployment is primarily attributed to assigning 4-bit quantization to components with fewer parameters and lower computational overhead, while more complex components are quantized at 6 or 8 bits. This outcome illustrates the inherent inefficiencies and deployment challenges associated with random quantization bitwidth assignments, underscoring the need for a more targeted approach in the selection process.
\vspace{-20pt}
\begin{table}[!htb]
\centering
\caption{Resource Utilization of Baseline 1}
\label{table:combinations_1}
\begin{tabular}{|c|c|c|c|c|c|c|}
\hline
\multirow{2}{*}{$n$} & \multirow{2}{*}{Combination} & \multicolumn{4}{c|}{Measured Utilization (\%)} & \multirow{2}{*}{RMSE} \\ \cline{3-6} 
&  & \multicolumn{1}{c|}{LUT} & \multicolumn{1}{c|}{DRAM} & \multicolumn{1}{c|}{BRAM}  & \multicolumn{1}{c|}{DSP} &  \\ \hline

12 & \multirow{3}{*}{4, 8, 8, 6, 8, 6, 8, 8, 4, 6}  & 95.0 & 92.3 &	100.0 & 100.0 &  4.46 \\ \cline{1-1} \cline{3-7} 
 18 & & \cellcolor{pink} \textbf{140.0} & \cellcolor{pink}\textbf{183.0} & \cellcolor{pink}\textbf{95.0} & \cellcolor{pink}\textbf{100.0} & \cellcolor{pink}\textbf{4.30} \\ \cline{1-1} \cline{3-7} 
 24 & & \cellcolor{pink} \textbf{143.4} & \cellcolor{pink}\textbf{191.3} & \cellcolor{pink}\textbf{95.0} & \cellcolor{pink}\textbf{100.0} & \cellcolor{pink}\textbf{4.20} \\ \hline

12 & \multirow{3}{*}{8, 8, 8, 4, 8, 6, 4, 4, 4, 4}   & 93.3 & 87.5 & 100.0	& 95.0 & 5.22 \\ \cline{1-1} \cline{3-7}
18 & &  \cellcolor{pink}\textbf{136.5} & \cellcolor{pink}\textbf{172.8} &  \cellcolor{pink}\textbf{95.0} &  \cellcolor{pink}\textbf{95.0} &  \cellcolor{pink}\textbf{4.95} \\ \cline{1-1} \cline{3-7}
24 & &  \cellcolor{pink}\textbf{139.3} & \cellcolor{pink}\textbf{181.3} &  \cellcolor{pink}\textbf{95.0} &  \cellcolor{pink}\textbf{100.0} & \cellcolor{pink} \textbf{5.24} \\ \hline

12 & \multirow{3}{*}{4, 8, 8, 4, 8, 4, 8, 8, 8, 8}  & 86.2 & 90.0 &	100.0 &	95.0 & 4.50 \\ \cline{1-1} \cline{3-7}
18 & &  \cellcolor{pink}\textbf{115.9} &  \cellcolor{pink}\textbf{162.0} &  \cellcolor{pink}\textbf{100.0}	&  \cellcolor{pink}\textbf{95.0} & \cellcolor{pink}\textbf{4.26} \\ \cline{1-1} \cline{3-7}
24 & & \cellcolor{pink} \textbf{124.8} &  \cellcolor{pink}\textbf{185.5} &  \cellcolor{pink}\textbf{100.0}	&  \cellcolor{pink}\textbf{95.0} & \cellcolor{pink}\textbf{4.33}  \\ \hline

12 & \multirow{3}{*}{6, 4, 6, 8, 6, 4, 4, 8, 6, 6}  & 72.9 & 73.7 & 100.0 & 85.0 & 4.49 \\ \cline{1-1} \cline{3-7}
18 & & \cellcolor{pink}\textbf{94.51} &  \cellcolor{pink}\textbf{121.7} &  \cellcolor{pink}\textbf{100.0}	& \cellcolor{pink} \textbf{87.5} &  \cellcolor{pink}\textbf{4.59} \\ \cline{1-1} \cline{3-7}
24 & & \cellcolor{pink}\textbf{102.5} &  \cellcolor{pink}\textbf{142.3} &  \cellcolor{pink}\textbf{100.0}	& \cellcolor{pink} \textbf{85.0} & \cellcolor{pink} \textbf{4.84} \\ \hline

12 & \multirow{3}{*}{8, 4, 6, 6, 4, 6, 4, 8, 4, 8}  & 79.1 & 70.2 &	100.0 & 80.0 & 4.93 \\ \cline{1-1} \cline{3-7} 
 18 & & \cellcolor{pink}\textbf{113.4} & \cellcolor{pink}\textbf{138.2} & \cellcolor{pink} \textbf{95.0} & \cellcolor{pink} \textbf{82.5} & \cellcolor{pink} \textbf{4.67} \\ \cline{1-1} \cline{3-7} 
 24 & & \cellcolor{pink}\textbf{116.0} & \cellcolor{pink}\textbf{144.0} & \cellcolor{pink} \textbf{95.0} & \cellcolor{pink} \textbf{80.0} & \cellcolor{pink} \textbf{4.35} \\ \hline

\end{tabular}
\end{table}
\vspace{-22pt}

\subsubsection{Baseline 2: Experience-Based Combination}
\label{subsubsec:baseline2}

Building upon insights garnered from Baseline 1, Baseline 2 refined the quantization by specifically assigning a 4-bit quantization to the MHA and FFN modules, which are characterized by higher parameter counts and computional overhead. Subsequently, these revised configurations underwent synthesis and were appraised using the identical evaluative benchmarks established in Baseline 1. Table \ref{table:combinations_2} presents the outcomes of Baseline 2, illustrating the successful FPGA deployment of all revised model configurations, thereby validating the efficacy of targeted bitwidth assignments. 

Additionally, the latter two columns of Table \ref{table:combinations_2} detail the models' precision in terms of RMSE, alongside comparisons to Baseline 1. We found that while 7 out of the 15 models demonstrated a decrease in RMSE, the remaining 8 models experienced an increase in RMSE. Among these 7 models, 4 models had already implemented a 4-bit quantized FFN during Baseline 1, necessitating only the quantization of MHA to 4-bit in Baseline 2. It is posited that the imposition of a 4-bit constraint on the FFN module may act as a computational bottleneck. However, the concurrent reduction of MHA's bitwidth did not substantially exacerbate RMSE outcomes. Conversely, the additional reduction in bitwidths for FFN and MHA for the remaining three models potentially alleviated overfitting, thereby enhancing the models' generalization capability.

\vspace{-15pt}
\begin{table}[!htb]
\centering
\caption{Resource Utilization of Baseline 2}
\label{table:combinations_2}
\begin{tabular}{|c|c|c|c|c|c|c|c|}
\hline
\multirow{2}{*}{$n$} & \multirow{2}{*}{Combination} & \multicolumn{4}{c|}{Measured Utilization (\%)} & \multicolumn{2}{c|}{RMSE} \\ \cline{3-8} 
&  & \multicolumn{1}{c|}{LUT} & \multicolumn{1}{c|}{DRAM} & \multicolumn{1}{c|}{BRAM}  & \multicolumn{1}{c|}{DSP} & Value & Change \\ \hline

12 & \multirow{3}{*}{4, 8, \underline{4}, 6, 8, \underline{4}, 8, 8, 4, 6}  & 62.2 & 27.3 &	100.0 & 90.0 & 4.78 & $\uparrow$7.17\% \\ \cline{1-1} \cline{3-8} 
 18 & & 76.8 &	61.3 & 100.0 & 85.0 & 4.71 & $\uparrow$9.53\% \\ \cline{1-1} \cline{3-8}  
 24 &  & 81.2 &	72.8 & 100.0 & 90.0 & 4.29 & $\uparrow$2.14\%\\ \hline

12 & \multirow{3}{*}{8, 8, \underline{4}, 4, 8, \underline{4}, 4, 4, 4, 4}   & 66.0 & 38.8 & 100.0	& 87.5 &  5.24 & $\uparrow$0.38\%\\ \cline{1-1} \cline{3-8} 
18 & & 80.5 & 68.8 & 100.0 & 85.0 & 4.87 & $\downarrow$1.62\% \\ \cline{1-1} \cline{3-8} 
24 & & 81.9 & 72.8 & 100.0 & 85.0 & 4.98 & $\downarrow$4.96\% \\ \hline

12 & \multirow{3}{*}{4, 8, \underline{4}, 4, 8, \underline{4}$^*$, 8, 8, 8, 8}  & 63.3 & 26.7 &	100.0 &	85.0 & 4.47 & $\downarrow$0.67\%\\ \cline{1-1} \cline{3-8} 
18 & & 75.1 & 53.7 & 100.0 & 85.0 &  4.47 & $\uparrow$4.93\%\\ \cline{1-1} \cline{3-8} 
24 & & 80.5 & 65.2 & 100.0 & 85.0 &  4.30 & $\downarrow$0.69\%\\ \hline

12 & \multirow{3}{*}{6, 4, \underline{4}, 8, 6, \underline{4}$^*$, 4, 8, 6, 6}  & 59.7 & 24.8 &	100.0 & 80.0 & 4.72 & $\uparrow$5.12\% \\ \cline{1-1} \cline{3-8} 
18 & & 71.7 & 48.2 & 100.0 & 77.5 &  4.48 & $\downarrow$2.40\%\\ \cline{1-1} \cline{3-8} 
24 & & 76.2 & 59.5 & 100.0 & 82.5 &  4.44 & $\downarrow$8.26\% \\ \hline

12 & \multirow{3}{*}{8, 4, \underline{4}, 6, 4, \underline{4}, 4, 8, 4, 8}  & 62.9 & 27.0 &	100.0 & 70.0 & 4.97 & $\uparrow$0.81\% \\ \cline{1-1} \cline{3-8}  
 18 & & 77.4 & 61.0 & 100.0 & 70.0 & 4.65 & $\downarrow$0.43\% \\ \cline{1-1} \cline{3-8}  
 24 & & 79.3 & 65.0 & 100.0 & 70.0 & 4.46 & $\uparrow$2.53\% \\ \hline
\multicolumn{8}{l}{\small $^*$ FFN was already quantized to 4-bit in Baseline 1.} 
\end{tabular}
\vspace{-15pt}
\end{table}

\subsubsection{Our approach}
\label{subsubsec:our_aproach}

Baseline 1 demonstrated the pitfalls of relying solely on random configuration selections, often culminating in inefficient and unsuccessful deployments. Moreover, Baseline 2 incorporated empirical insights to enhance deployment success rates. However, this approach did not stabilize model accuracy, as evidenced by unpredictable fluctuations in RMSE. 

In response, our study implemented the Algorithm~\ref{alg:resource_aware_quantization} detailed in Section \ref{subsec:res-filter}. We applied it across three distinct input sequence lengths ($n$), each exploring a comprehensive range of 59,049 possible quantization combinations, referred to as $C_\text{all}$. To streamline this vast array, specific resource utilization thresholds were set: $T_\text{LUTs}\!=\! 80\%$, $T_\text{DRAM}\!=\!100\%$, $T_\text{BRAM}\!=\!100\%$, and $T_\text{DSPs}\!=\!100\%$.

\begin{figure}[!htb]
    \vspace{-10pt}
    \centering
    \includegraphics[width=.85\textwidth]{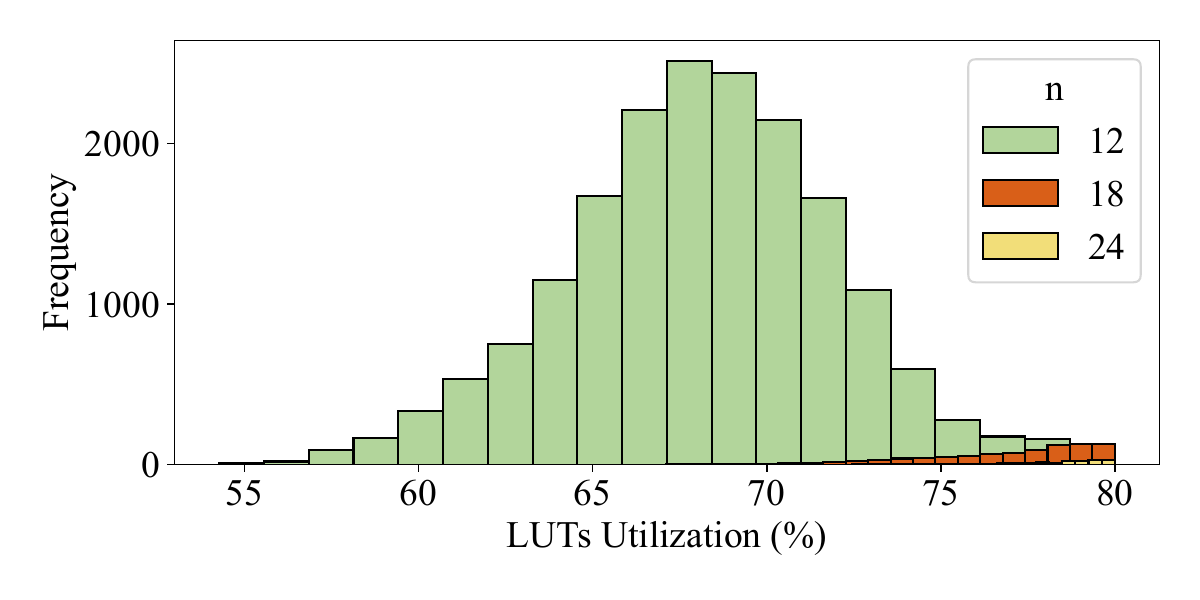}
    \caption{Distribution of Candidate Mixed-Precision Bitwidth Combinations}
    \label{fig:candidate_distribution}
    \vspace{-10pt}
\end{figure}

These thresholds effectively narrowed the feasible combinations by 69.3\% for $n\!=\!12$, 98.5\% for $n\!=\!18$, and 99.7\% for $n\!=\!24$. The histogram in Figure \ref{fig:candidate_distribution} illustrates the distribution of remaining quantization combinations across different input sequence lengths after applying the specified resource utilization thresholds. Notably, the majority of combinations for $n\!=\!12$ fall within the middle range of LUT utilization, while the distributions for $n\!=\!18$ and $n\!=\!24$ are significantly narrower, reflecting the stringent filtering process that effectively reduces the number of viable configurations, especially for higher sequence lengths.

After score-based filtering, 5 candidates for each input sequence length were subjected to training and evaluation, with their performance and resource utilization detailed in Table \ref{table:combinations_3}. This table presents ``Estimated Utilization (\%)'' derived from our knowledge database and ``Measured Utilization (\%)'' obtained from Vivado synthesis, affirming the precision and reliability of our database. For example, when $n\!=\!12$, the estimated LUT utilization of 80.0\% closely mirrored the measured value of 85.7\%. It is noteworthy that the overhead from $O_\text{model}$, $O_\text{encoder layer}$, and $O_\text{middleware}$, typically accounting for 2.7\% to 4.0\% of LUT utilization, were excluded from these estimates. Including this overhead will bring the estimated results closer to the actual measurements.

\vspace{-20pt}
\begin{table}[!htbp]
\caption{Resource Utilization of Our Approach}
\label{table:combinations_3}
\centering
\scalebox{.92}{
\begin{tabular}{|c|c|c|c|c|c|c|c|c|c|c|}
\hline
\multirow{2}{*}{$n$} & \multirow{2}{*}{Combination} & \multicolumn{4}{c|}{Estimated Utilization (\%)$^*$} & \multicolumn{4}{c|}{Measured Utilization (\%)} & \multirow{2}{*}{RMSE} \\ \cline{3-10} 
&  & \multicolumn{1}{c|}{LUT} & \multicolumn{1}{c|}{DRAM} & \multicolumn{1}{c|}{BRAM}  & DSP & \multicolumn{1}{c|}{LUT} & \multicolumn{1}{c|}{DRAM} & \multicolumn{1}{c|}{BRAM} & DSP &  \\ \hline

\multirow{5}{*}{12} 
    & 6, 8, 6, 8, 6, 6, 8, 8, 8, 8 & 80.0  & 78.7 & 100.0 &	100.0  &  85.7 &	79.3	&100.0 & 100.0 & 3.81 \\ \cline{2-11} 
    & 8, 8, 6, 8, 8, 4, 8, 6, 8, 8 & 78.1  & 76.0 & 85.0  & 100.0 & 80.9 & 82.7	& 100.0 &	100.0  & 3.89  \\ \cline{2-11} 
    & 8, 8, 6, 8, 6, 4, 8, 8, 8, 8 & 78.0  & 76.0 &	85.0  & 100.0  & 80.3 &	80.0 &	100.0 & 100.0 & 3.77  \\ \cline{2-11} 
    & 8, 8, 4, 8, 8, 6, 8, 6, 8, 8 &  76.7 & 65.8 &	85.0  & 100.0  & 80.3 &	55.2 &	100.0 & 100.0 & 4.14  \\ \cline{2-11} 
    & 8, 8, 4, 8, 6, 6, 8, 8, 8, 8 &  76.6 & 65.8 &	85.0  & 100.0  & 79.2 &	52.5 &	100.0 & 100.0 & 4.07  \\ \hline

\multirow{5}{*}{18}                    
    &  8, 4, 4, 4, 4, 4, 8, 4, 8, 8  & 80.0 & 77.2 & 90.0 & 75.0 & 77.2 & 61.2 & 100.0 & 75.0 &  4.65 \\ \cline{2-11} 
    &  8, 4, 4, 4, 8, 4, 4, 4, 8, 8  & 79.8 & 77.2 & 90.0 & 70.0 & 77.8 & 61.2 & 100.0 & 75.0 &  5.16 \\ \cline{2-11} 
    &  8, 4, 4, 4, 4, 4, 4, 8, 8, 8  & 79.7 & 77.2 & 90.0 & 70.0 & 77.8 & 61.2 & 100.0 & 65.0 &  4.55 \\ \cline{2-11} 
    &  8, 6, 4, 4, 6, 4, 4, 4, 8, 8  & 79.7 & 74.5 & 85.0 & 75.0 & 76.3 & 56.5 & 100.0 & 80.0 &  4.19 \\ \cline{2-11} 
    &  6, 8, 4, 4, 6, 4, 4, 4, 8, 8  & 79.6 & 74.5 & 85.0 & 75.0 & 76.3 & 56.5 & 100.0 & 80.0 &  4.53 \\ \hline

\multirow{5}{*}{24} 
    & 6, 8, 4, 4, 4, 4, 4, 4, 8, 8 & 79.7 &	75.8 & 85.0 & 75.0 & 80.7 &	67.8 &	100.0 & 80.0 &  4.10 \\ \cline{2-11} 
    & 8, 6, 4, 4, 4, 4, 4, 4, 8, 6 & 79.9 &	75.7 & 85.0 & 75.0 & 80.4 &	67.7 &	100.0 &	80.0 &  4.49 \\ \cline{2-11} 
    & 8, 6, 4, 4, 4, 4, 4, 4, 6, 8 & 79.8 &	75.7 & 85.0 & 75.0 & 80.0 &	67.7 &	100.0 & 80.0 &  4.65 \\ \cline{2-11} 
    & 4, 6, 4, 4, 4, 4, 6, 4, 8, 8 & 79.6 &	78.5 & 85.0 & 85.0 & 77.7 &	67.8 &	100.0 & 75.0 &  4.23 \\ \cline{2-11} 
    & 6, 8, 4, 4, 4, 4, 4, 4, 8, 6 & 79.5 &	75.7 & 85.0 & 75.0 & 80.2 &	67.7 &	100.0 & 80.0 &  4.37 \\ \hline
\multicolumn{11}{l}{\small $^*$ The overhead from $O_\text{model}$, $O_\text{encoder layer}$, and $O_\text{middleware}$ is excluded.} 
\end{tabular}}
\vspace{-17pt}
\end{table}

Table \ref{table:combinations_3} also elucidates the impact of our method on test RMSE. Across different input sequence lengths, our approach consistently resulted in lower average RMSEs compared to Baseline 2. However, for an input sequence length of 18, while Baseline 1 showed a lower average RMSE than our approach, the models it produced were not deployable due to exceeding FPGA resource constraints. Notably, our approach often utilized 4-bit quantization for model components at $n\!=\!18$ and $n\!=\!24$, constrained by stringent resource thresholds. In contrast, Baseline 2 configurations typically used fewer than four components quantized at 4 bits. This observation suggests that refining our score-based filter to reduce reliance on 4-bit quantization could further improve RMSE, enhancing model performance across different sequence lengths while maintaining deployability.

\subsection{Experiment 3: Model Performance on Hardware}

Following the results of Experiment 2, accelerators associated with the lowest RMS for each input sequence length ($n$) underwent further performance evaluations using Vivado’s synthesis and actual hardware validation. Table \ref{tab:model_hardware} details the RMSE, inference time, power and energy consumption for each configuration.


\begin{table}[!htb]
\vspace{-15pt}
\footnotesize
\centering
\caption{Performance Comparison on Spartan-7 XC7S15 FPGA}
\label{tab:model_hardware}
\begin{tabular}{|c|c|c|c|c|c|c|c|}
\hline
\multirow{2}{*}{n} & \multirow{2}{*}{Combinations} & \multirow{2}{*}{RMSE} & \multirow{2}{*}{Time(ms)$^\dagger$} & \multicolumn{3}{c|}{Power(mW)$^\dagger$}  & \multirow{2}{*}{Energy(mJ)} \\ \cline{5-7}
& &  &  & \multicolumn{1}{c|}{Static} & Dynamic & Total & \\ \hline
12 & 8, 8, 6, 8, 6, 4, 8, 8, 8, 8 & 3.77  & 5.78 & 31 & 39 & 70 & 0.405 \\ \hline
18 & 8, 6, 4, 4, 6, 4, 4, 4, 8, 8 & 4.19  & 8.79 & 31 & 37 & 68 & 0.598 \\ \hline
24 & 6, 8, 4, 4, 4, 4, 4, 4, 8, 8 & 4.10  & 11.92 & 31 & 37 & 68 & 0.811 \\ \hline
\multicolumn{8}{l}{\small $\dagger$  The estimates obtained from GHDL and Vivado exhibit 2\% variance in time,} \\
\multicolumn{8}{l}{\small  \quad  and 5\% variance in power when compared to the actual hardware measurements.} 
\end{tabular}
\vspace{-15pt}
\end{table}

For an input sequence length of 12, deploying an accelerator configured with uniform 6-bit quantization was successfully implemented on the XC7S15 FPGA, as previously discussed in \cite{ling2024transformer}. This configuration achieved an RMSE of 3.76. In comparison, our mixed-precision quantized accelerator, as detailed in the first row of Table \ref{tab:model_hardware}, registered a slightly higher RMSE of 3.77. The inclusion of 8-bit quantized components restricted the maximum clock frequency to 100 MHz, compared to 125 MHz for the uniform 6-bit configuration. This contributed to a bottleneck that increased the inference time for our mixed-precision quantized accelerator by 25\%, resulting in a duration of 5.78 ms. Although there was a 12.85\% reduction in power consumption, the extended inference time led to an increase in energy consumption to 0.405 mJ, illustrating that our mixed-precision quantized accelerator underperformed relative to the uniform 6-bit quantized accelerator. However, we believe that optimizing the score-based filter to limit mixed-precision combinations to 4 and 6-bit could potentially alleviate the maximum clock frequency constraints imposed by 8-bit components, potentially reducing inference time and decreasing energy consumption.

For $n=18$, prior implementations were limited to uniform 4-bit quantization, resulting in poor model precision with an RMSE of 5.286. Our mixed-precision quantized accelerator significantly enhanced model precision, reducing the RMSE to 4.19—an improvement of 20.73\%, as detailed in the second row of Table \ref{tab:model_hardware}. Moreover, compared to the floating-point counterpart, our mixed-precision quantized accelerator recorded only a 3.33\% higher RMSE. For an input sequence length of 24, previous efforts could not deploy any accelerators even with uniform 4-bit quantization due to excessive resource demands. In this study, however, adaptive memory allocation enabled the deployment of a uniform 4-bit quantized accelerator, as outlined in Section \ref{subsec:experiment1}. Our mixed-precision quantized accelerator further improved upon this, reducing the RMSE by 24.85\%. Relative to the floating-point counterpart, our mixed-precision quantized accelerator showed only a 2.80\% higher RMSE.

However, increasing the input sequence length from 12 to 24 nearly doubled the inference time. Notably, dynamic power consumption was reduced by 2 mW, attributable to a 20\% decrease in DSPs utilization, as detailed in Table \ref{table:combinations_3}. Despite this reduction in power consumption, the prolonged inference duration led to a doubling of the energy required per inference. Furthermore, extending the input sequence length did not reduce RMSE for the \textit{AirU} dataset, indicating that the configuration detailed in the first row of Table \ref{tab:model_hardware} is the most effective for this particular dataset. This outcome underscores the complex interplay between model architecture, quantization strategy, and dataset characteristics. Despite these nuances, our primary contribution is the enhanced deployability of sophisticated Transformer models on resource-constrained devices. We anticipate that further validation across different datasets will substantiate the broader applicability and benefits of our approach.


\section{Related Work}
\label{sec:related_work}

Recent research has increasingly focused on optimizing Transformers through quantization to balance computational efficiency with model precision~\cite{chitty2023survey}. Zhang et al.~\cite{zhang2024large} reviewed vector quantization and K-Means-based methods applied to large language models for time-series analysis. In addition, Zhao et al.~\cite{zhao2024sparse} proposed an innovative Transformer architecture that replaces the traditional FFN module with vector quantization to enhance efficiency and precision in multivariate time-series forecasting. However, the substantial computational resources required for their implementations, as evidenced by their use of the NVIDIA A100 80GB GPU, highlight the challenges of deploying such models on IoT devices.

In the realm of FPGA-targeted deployment, accelerating quantized Transformers presents specific challenges due to the stringent resource limitations of these platforms. Okubo et al.~\cite{okubo2024cost} developed a compact Transformer model that incorporates Neural Ordinary Differential Equations, significantly reducing both parameter count and resource utilization. This model has shown considerable gains in inference speed and energy efficiency on a modest-sized ZCU104 FPGA. Furthermore, the application of mixed-precision quantization, which assigns different quantization bitwidths to various model components, is increasingly acknowledged as essential for optimizing FPGA deployment~\cite{rakka2024review}. For example, Chang et al.~\cite{chang2021rmsmp} demonstrated that their mixed-precision quantization approach, which assigns multiple precision levels within the weight matrix at the row level, enhances the efficiency of quantized BERT models on System-on-Chip-based FPGAs (ZYNQ XC7Z020 and XC7Z045). Additionally, Li et al.~\cite{li2022auto} have implemented mixed-precision quantized Vision Transformer accelerators on the ZCU102 FPGA, utilizing distinct bitwidths for model weights and activations. 

However, it is critical to recognize that the FPGAs used in these studies are relatively large, and the FPGA resource capacities often exceed those required for typical applications. This underscores a gap in research for deploying Transformer-based models on embedded FPGAs, where resource constraints are more pronounced. Our study addresses this gap by proposing adaptive resource allocation and resource-aware mixed-precision quantization to enhance the deployability of Transformer models on FPGAs with limited resources. Additionally, this research offers a detailed analysis of the Transformer accelerators, specifically tailored for time-series forecasting, thereby advancing practical understanding of efficient model deployment on resource-constrained platforms.

\section{Conclusion and Future Work}
\label{sec:conclusion_future}

This study has demonstrated the efficacy of adaptive resource allocation and mixed-precision quantization for deploying Transformer models for time-series forecasting on embedded FPGAs. Our approach, underpinned by a comprehensive knowledge database for precise resource estimation, facilitates the effective implementation of mixed-precision quantization. By integrating resource awareness into our workflow, we have streamlined the deployment process on resource-constrained devices, ensuring both deployability and maintaining competitive model precision. This study sets a benchmark for future optimizations in similar applications in ubiquitous computing.

In the future, we plan to refine our score-based filter to enhance the precision and deployment efficiency of the quantized Transformer model. Additionally, we aim to validate our approach across other datasets, verifying its advantages and applicability in diverse scenarios. Furthermore, we intend to explore the potential of mixed-scheme quantization strategies, which combine different quantization techniques within a single model architecture, aiming to further optimize computational efficiency.

\section*{Acknowledgments} 
The authors gratefully acknowledge the financial support provided by the Federal Ministry for Economic Affairs and Climate Action of Germany for the RIWWER project (01MD22007C). 
\bibliographystyle{splncs04_unsort}
\bibliography{reference}
\end{document}